\documentclass[dvipsnames,format=sigconf]{acmart}

\usepackage[ruled,linesnumbered]{algorithm2e}
\usepackage{bbold}
\usepackage{subcaption}

\AtBeginDocument{%
  \providecommand\BibTeX{{%
    \normalfont B\kern-0.5em{\scshape i\kern-0.25em b}\kern-0.8em\TeX}}}

\setcopyright{acmcopyright}
\copyrightyear{2023}
\acmYear{2023}
\acmDOI{XXXXXXX.XXXXXXX}

\acmConference[GECCO '23]{The Genetic and Evolutionary Computation Conference}{July 15--19,
  2023}{Lisbon, Portugal}
\acmPrice{15.00}
\acmISBN{978-1-4503-XXXX-X/18/06}

\begin{document}

\title{Towards Large-Scale Simulations of Open-Ended Evolution in Continuous Cellular Automata}

\author{Bert Wang-Chak Chan}
\affiliation{%
  \institution{Google Research, Brain team}
  \city{Tokyo}
  \country{Japan}}
\email{bertchan@google.com}

\renewcommand{\shortauthors}{Chan}

\begin{abstract}
Inspired by biological and cultural evolution, there have been many attempts to explore and elucidate the necessary conditions for open-endedness in artificial intelligence and artificial life. Using a continuous cellular automata called Lenia as the base system, we built large-scale evolutionary simulations using parallel computing framework JAX, in order to achieve the goal of never-ending evolution of self-organizing patterns. We report a number of system design choices, including (1) implicit implementation of genetic operators, such as reproduction by pattern self-replication, and selection by differential existential success; (2) localization of genetic information; and (3) algorithms for dynamically maintenance of the localized genotypes and translation to phenotypes. Simulation results tend to go through a phase of diversity and creativity, gradually converge to domination by fast expanding patterns, presumably a optimal solution under the current design. Based on our experimentation, we propose several factors that may further facilitate open-ended evolution, such as virtual environment design, mass conservation, and energy constraints.
\end{abstract}



\graphicspath{ {./figures/} }
\begin{teaserfigure}
  \includegraphics[width=\textwidth]{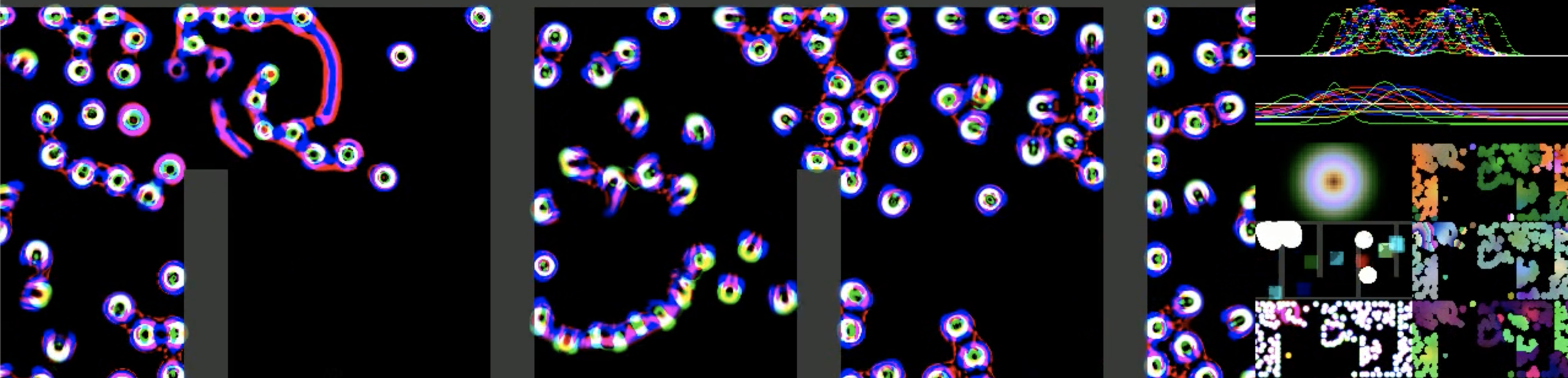}
  \caption{Evolutionary simulation using continuous cellular automata.}
  \label{fig:teaser}
\end{teaserfigure}

\received{10 February 2023}

\maketitle

\section{Introduction}

\subsection{Open-endedness}
One of the grand challenges in artificial intelligence and artificial life is \textit{open-endedness} (OE) \cite{Stanley2017-le}, that is, the hypothetical ability of an evolving or learning system to continuously produce novelty and/or increasing complexity.

In the field of artificial life (ALife), OE is called \textit{open-ended evolution} (OEE). This field of research is inspired by the evolution of biological life, which is able to produce lifeforms of seemingly infinite diversity in multiple levels of organizational hierarchy. ALife researchers wish to replicate this property in artificial systems, like computer simulation and artificial chemistry \cite{Soros2014-vy, Sayama2011-wi, Sayama2018-oc, Duim2017-ht}, and have been discussing the necessary conditions for OEE \cite{Banzhaf2016-kt, Taylor2019-xc, Taylor2016-pf, Packard2019-qt, Packard2019-ak}.

In artificial intelligence (AI), this topic is simply refer to as open-endedness. Researchers are fascinated by the apparent unlimited capacities of the human mind, like learning new skills, crafting new artefacts, or pursuing new goals, all empowered by our use of imagination and language. This is especially powerful in the collective sense, producing science, literature, cultures, and civilizations.

In this paper, we report the first steps in an on-going study of OEE in an ALife system called Lenia. Using large-scale simulations with parallel computing, we hope to elucidate the requirements for OEE through experimentation and model iteration, and to understand the essences and limitations of biological and computational evolutionary processes.

\subsection{Continuous cellular automata}
The target system is a family of continuous cellular automata (continuous CA) called Lenia \cite{Chan2019-po, Chan2020-iu}, originated from Conway's Game of Life, but with continuous states, space and time instead of discrete ones. We choose this base system because it is highly expressive and flexible, able to generate high diversity of dynamical patterns as well as emergent phenomena like directional movement, self-repair, self-defence, attraction / repulsion, self-replication, small pattern emission, metamorphosis, differentiation, swarming, etc. The gradual reveal of these phenomena one-by-one from previous studies of Lenia is not unlike life on Earth that undergone major evolutionary transitions like endosymbiosis and multicellularity.

Like other ALife system design, there is a duality of genotype and phenotype in Lenia. Lenia can be highly parameterized, depending on which variation of the system is used, the number of model parameters (known as \textit{genes}) ranges from 2 to hundreds. For a particular set of parameter values (\textit{genotype}), there can exist patterns (\textit{phenotype}) which are self-organizing dynamical structures of non-zero states.

These patterns can be spatially localized, which are called ``virtual creatures'', not only because they resemble biological living creatures, they also possess \textit{enactivist agency} \cite{Barandiaran2017-sc, Hamon2022-od}, meaning that they are able to perceive the virtual environment inside the CA, and react accordingly, like repel or attract another pattern, chase after it, or metamorphose into an another phenotype. These enactivist agents are constantly under precarious conditions, easily destabilized by themselves or by the environment. Only a subset of these agents, usually after being evolved by evolutionary algorithms or trained by machine learning, are self-sufficient and robust from perturbations. They are the basic units of selection in our evolutionary algorithm.

\subsection{Evolutionary algorithm}
In the previous works of controlled evolution of virtual creatures in Lenia, usually there is an inner loop of CA simulation, combined with an outer loop of genetic operators. These genetic operators are carried out by hand or by automation: inheritance and reproduction by saving the genotype and phenotype data inside disk or memory, retrieving them when needed; mutation of genotype by random incremental changes; and selection by a combination of the following criteria: survival (not vanishing nor turning into an exploding pattern after a number of simulation steps), optimizing a fitness function (e.g. maximizing the pattern's travelling speed), novelty (unexplored inside a behavioral space or parameter space), subjective preferences (e.g. novelty, complexity, aesthetics or ``interestingness'').

\begin{figure}[htbp]
\centering
\includegraphics[width=1.1\linewidth]{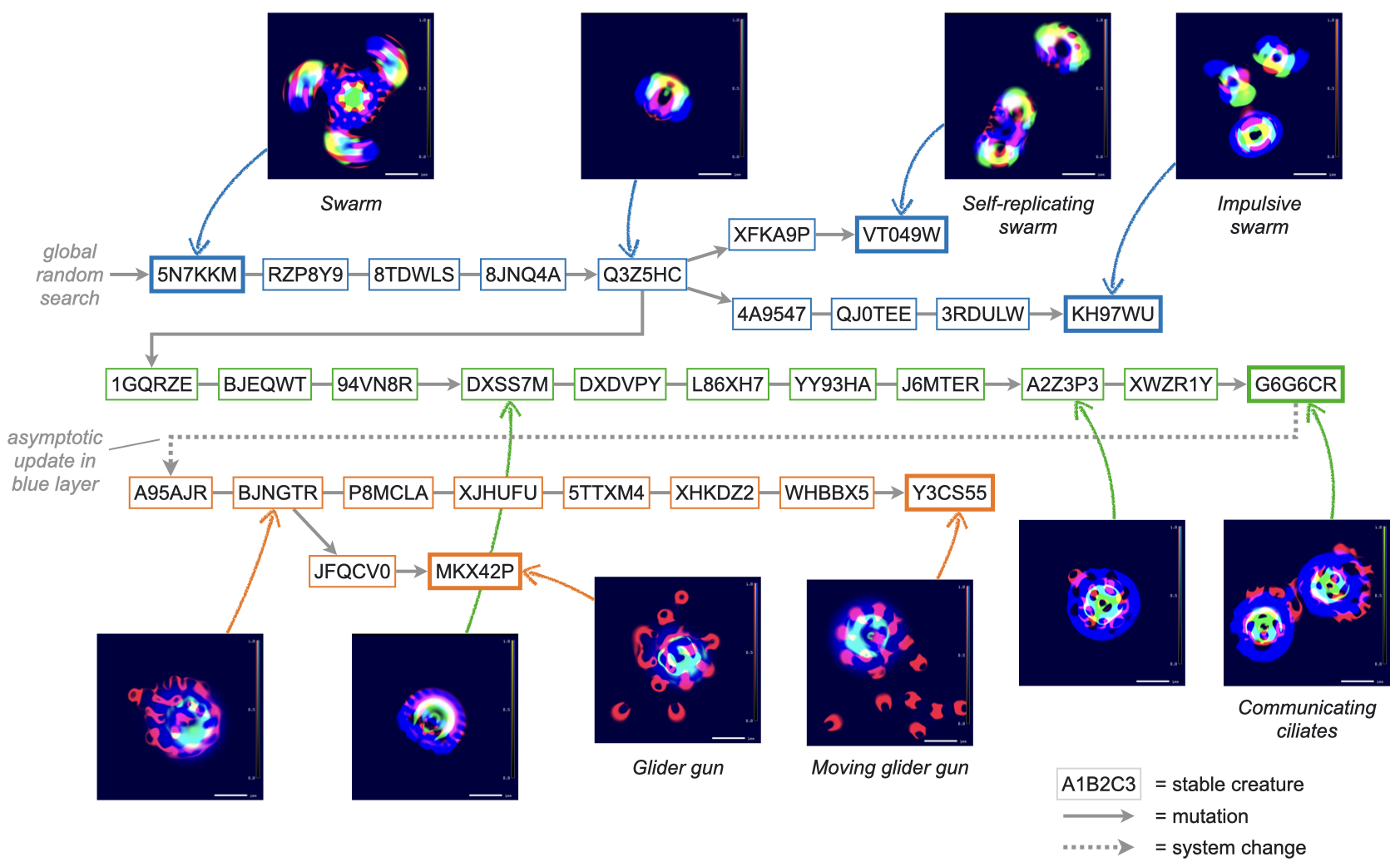}
\caption{The ``\textit{Aquarium} lineage'', a series of Lenia species previously obtained by controlled evolutionary algorithm, many demonstrate emergent novel behaviors. This work aims to replicate these emergence through large-scale intrinsic evolution.}
\label{fig:lineage} 
\end{figure}

Under this paradigm, one is able to discover remarkable patterns with \textit{transformational novelty} \cite{Taylor2019-xc, Banzhaf2016-kt}. A notable example is a series of patterns called the ``\textit{Aquarium} lineage'' (Figure~\ref{fig:lineage}). Starting from a new species called \textit{Aquarium} with a randomly generated genotype, after a long history of mutations and occasional system-wise change (at one point a channel was changed to asymptotic update \cite{Kawaguchi2021-gl}), the evolutionary process produced virtual creatures with exceptional capabilities, including swarming, pattern emission, differentiation, and certain behaviors that are difficult to be characterized (cell-cell ``communication'' and temporary ``coagulation'', resembling biological immune systems).

This approach is akin to preparing ``Petri dishes'' of small-scale replicable CA simulations, with a ``laboratory'' of scientists or machines applying the genetic operators. However, biological evolution does not occur in this kind of controlled lab-like environment, but instead inside a vast ocean of infinite possibilities, where creature-creature, creature-environment, group-group and species-species interactions occur continuously. This would be a better if not the only way to obtain highly complex structures and highly sophisticated interactive behaviors.

Instead of explicit implementation of genetic operators in an outer loop, in this work we aim to design the system such that they are implicitly implied or spontaneously emerge inside the simulation. Inheritance is realized by localized genotypes and phenotypes; reproduction relies on the emergence of pattern self-replication; selection is performed by the emergent \textit{differential existential success} (i.e. ability to self-preserve and survive the competition for space). This implicit approach of \textit{intrinsic evolution} may increase the chance of OEE \cite{Banzhaf2016-kt}, but with drawbacks like the prerequisite of self-replicating patterns, non-trivial ways to encourage or penalize certain behaviors, and complicated implementation details. In later parts of this paper, we will demonstrate our attempts to mitigate these drawbacks.

\subsection{Large-scale simulations}
Another kind of difficulty for intrinsic evolution is the need of large-scale simulation in terms of space and time. The world must be big enough to allow a large number of virtual creature interactions as well as \textit{in situ} selection to occur. Simulation time need to be much longer for the observation of new emergent phenomena, and at least 20 frames-per-second to allow real-time observation. In this regard, we adopt the recent developments in parallel computing, namely the JAX framework \cite{Frostig2018-jw} in Python that automatically utilize GPU or TPU resources.

We designed and implemented new algorithms that enable the localization of genetic information, such that different species can co-exist in the same simulation, species-species interactions and pixel-level mutations become possible. Algorithms were also designed to recognize and penalize any exploding patterns that would quickly expanding all over the world and wipe out all other virtual creatures.

The goal is to design an intrinsic evolution framework that can replicate the successful \textit{Aquarium} lineage and beyond. If multiple transformative novelties can emerge in a single run of such system, we can be highly confident that OEE can be achieved inside continuous cellular automata.

\section{Methods}

\subsection{The target system}
Modified from the famous CA Conway's Game of Life, the continuous CA Lenia consists of a grid of cells and a set of local update rules, with many of the discrete aspects in the Game of Life turned into continuous ones.
Here we use a 2-dimensional, 3-channel, 15-kernel version of the original Lenia.

\begin{itemize}
    \item A large size gird world of $512 \times 1024$ pixels, with repeating toroidal boundaries.
    \item 3 channels ($n_c = 3$), meaning each cell has a vector state of 3 values within the unit range.
    \item 15 kernels, with 3 self-connecting kernels for each channel ($n_\text{self} = 3$) and 2 cross-connecting kernels for each pair of channels ($n_\text{cross} = 2$), i.e. $n_k = n_\text{self} \, n_c + n_\text{cross} \, {n_c \choose 2} = 15$.
    \item Each kernel consists of two Gaussian bumps with 6 parameters (genes): center, width, height of each bump ($\mathsf{r_1, w_1, b_1}$, $\mathsf{r_2, w_2, b_2}$).
    \item Each kernel is paired with a growth mapping, which is a Gaussian function with 3 parameters (genes): center, width, height ($\mathsf{m, s, h}$).
    \item The original Lenia update rule: $A \mapsto [A + G(\tilde K*A)]_0^1$.
\end{itemize}

The original Lenia algorithm (Algorithm~ \ref{alg:extrinsic}) consists of an inner loop for CA simulation and an outer loop for controlled evolution. The new algorithm with intrinsic evolution (Algorithm~\ref{alg:intrinsic}) only consists of a single loop, integrating both simulation and evolution in one iteration.

\begin{algorithm}
\caption{Lenia with Controlled Evolution}
\label{alg:extrinsic}
add a seed pattern into the gene pool\;
\tcp{evolution loop}
\Repeat{reaching the target number of evolution steps}{
select a pattern from the gene pool according to a set of criteria, load its genotype and phenotype\;
mutate the genotype\;
initialize the phenospace tensor $A$ with the phenotype\;
initialize global parameters with the genotype\;
pre-calculate normalized kernels $\tilde K = K / (K*\mathbb 1)$\;
\tcp{simulation loop}
\Repeat{reaching the target number of simulation steps}{
    convolve the phenotype with kernels $\tilde K*A$\;
    apply growth mapping and incremental update $A_i \mapsto [A_i + \sum_k G_k(\tilde K_k*A_j)]_0^1$\;
    display or record the pattern if needed\;
}
save the genotype and the resulting phenotype into the gene pool\;
}
\end{algorithm}

\begin{algorithm}
\caption{Lenia with Intrinsic Evolution}
\label{alg:intrinsic}
initialize the phenospace tensor $A$ with seed phenotype\;
initialize the genospace tensor $P$ with seed genotype\;
pre-calculate kernel rings $K_r$ and size $K_r * \mathbb 1$\;
pre-calculate kernel for genospace diffusion $K_\text{diff}$\;
pre-calculate kernel for penalization $K_\text{pen}$\;
\tcp{simulation and evolution loop}
\Repeat{reaching the target number of simulation steps}{
    add environmental elements to the phenospace\;
    diffuse the genospace\;
    convolve the phenospace with every kernel ring $K_r*A$\;
    sum convolutions weighted by kernel function $(K*A) / (K*\mathbb 1)$\;
    calculate the alpha channel $\alpha$\;
    mask the genospace by alpha channel\;
    get the penalizing area of the phenospace  $E_\text{pen}$\;
    apply growth mapping and incremental update $A_i \mapsto \big[A_i + \sum_k G_k\big((K_k*A_j) / (K_k*\mathbb 1)\big)\big]_0^1$\;
    mutate the genospace\;
    penalize exploding patterns\;
    display or record the pattern if needed\;
}
\end{algorithm}

Note: $[x]_a^b = \max(\min(x, b), a)$ is the clipping function; $K*\mathbb 1 = \sum_{xy} K$ is the sum of the kernel used as a normalizing factor. Any calculation of convolution, including $\tilde K*A$ here, can be sped up by using the convolution theorem and fast Fourier transform (FFT):
\[X*Y = \mathscr{F}^{-1}(\mathscr{F} X \odot \mathscr{F} Y)\]

\subsection{Localizing genotypes} \label{seg:localize}
The genetic information is stored inside a 4-tensor $P \in \mathbb{R}^{p k x y}$, where $p$ is the parameter types ($\mathsf{r_1, w_1, b_1}$, $\mathsf{r_2, w_2, b_2}$, $\mathsf{m, s, h}$), $k$ is the kernel index ($0 \ldots 14$), $x$ and $y$ are the spatial dimensions. We call this tensor the \textit{genospace} in which parameter values are stored, in contrast to the \textit{phenospace} (i.e. the world) $A \in \mathbb{R}^{c x y}$ where cell states are stored ($c$ is the channel index $0 \ldots 2$). The genospace is initialized by filling the genome of each seed pattern near its initial occupying area.

First define Boolean masks $\Delta_\text{all}(T)$, $\Delta_\text{sum}(T)$, and $\Delta_\text{same}(T)$, that are flattened by logical `and' operator, flattened by sum, and dimension-preserving, respectively.
\[\Delta_\text{all}(T) = \big\{(\chi) | T(\varphi,\chi) > \epsilon, \forall \varphi\big\}\]
\[\Delta_\text{sum}(T) = \big\{(\chi) | {\textstyle \sum_\varphi} T(\chi) > \epsilon\big\}\]
\[\Delta_\text{same}(T) = \big\{(\varphi,\chi) | T(\varphi,\chi) > \epsilon\big\} \]
where $\chi$ and $\varphi$ are the lists of all spatial dimensions (e.g. $x,y$) and all non-spatial dimensions (e.g. $p,k,c$) of tensor $T$, respectively; $\epsilon$ is a small cutoff threshold e.g. $\epsilon = 0.01$.

In each simulation iteration, the genospace is masked by an alpha channel, like in Neural CA \cite{Mordvintsev2020-dv}. The alpha channel mask is calculated as nonzero portions of potential $U = G\big( (K*A) / (K*\mathbb 1) \big)$, used as a proxy of the envelope of creatures in the phenospace (cf. \cite{Sayama2018-oc} uses a blur operation instead).
\[P \mapsto P \odot \alpha\]
\[\alpha = \Delta_\text{all}(U) \wedge \Delta_\text{all}(P)\]

\subsection{Diffusing genotypes} \label{sec:diffuse}
In each simulation iteration, non-zero portions of the genospace is diffused to its vicinity, preparing for any directional movements of the creatures. This operation is a convolution with a diffusion kernel, then divided by its volume. The diffusion kernel $K_\text{diff}$ is a simple disk of radius $r_\text{diff}$, pre-calculated during initialization.
\[P \mapsto \left[\frac{P*K_\text{diff}}{\Delta_\text{same}(P)*K_\text{diff}}\right]_0^1\]


This diffusion operation also defines how the genospace behaves when areas of different genotypes collide. According to the current design, they diffuse slowly with each other and will eventually settle down into an equilibrium of average values.

\subsection{Translating localized genotypes to phenotypes}
It is trivial to utilize the localized genetic information for the growth function $G(x; \mathsf{m,s,h}) = \mathsf h \exp(-(x-\mathsf m)^2 / 2 \mathsf s^2)$, simply using components $\mathsf{m, s, h}$ from the genospace tensor $P$ as its parameters values for a particular kernel $k$ and spatial location $(x,y)$. 
\[G\big(K_k*A(x,y); P_{\mathsf mkxy}, P_{\mathsf skxy}, P_{\mathsf hkxy}\big)\]

However, for the kernels, since they are no longer global but change over space, we cannot use the convolution theorem to speed up calculation. An alternative is to use plain convolution, but the kernels (large matrices of convolution weights) need to be recalculated for each pixel, this will significantly slow down the simulation. Inspired by the discrete ring-like kernels in MNCA \cite{Kraakman2021-mj}, the solution proposed here is to separate a kernel into multiple rings.

Instead of pre-calculating (``casting'' or ``molding'') the kernels before use, the kernel matrices can be dissembled into a set of discrete rings. When used per pixel, the rings are convolved with the phenospace using the convolution theorem, and recombined using the kernel function $\mathscr k(x; \mathsf{r,w,b}) = \mathsf b \exp(-(x-\mathsf r)^2 / 2 \mathsf w^2)$ as weights. This is a good approximation of the original kernels in low spatial resolutions. In our implementation, the kernel of size $R = 12$ is divided into 24 rings ($n_\text{ring} = 24, r = 0 \ldots 23$), each has weight $\mathscr k_r$ by combining values of two kernel functions ($i = 0 \ldots 1$).
\[\mathscr k_r = \sum_i \mathscr k\big(r/n_\text{ring}; P_{\mathsf r_ikxy}, P_{\mathsf w_ikxy}, P_{\mathsf b_ikxy}\big)\]
\[K \approx \sum_r \mathscr k_r K_r\]
\[\frac{K*A}{K*\mathbb 1} \approx \frac{(\sum_r \mathscr k_r K_r)*A}{(\sum_r \mathscr k_r K_r)*\mathbb 1} = \frac{\sum_r \mathscr k_r (K_r*A)}{\sum_r \mathscr k_r (K_r*\mathbb 1)} \]
\[K_r*A = \mathscr{F}^{-1}(\mathscr{F} K_r \odot \mathscr{F} A)\]

The kernel rings are the same over space, therefore the tensors $K_r$, their Fourier transform $\mathscr{F} K_r$, and the normalizing factors $K_r*\mathbb 1 = \sum_{xy} K_r$ can be pre-calculated during initialization.

Special testing and calibration have been done to check any discrepancy between the original ``cast'' kernels and the recombined ring-wise kernels, especially any undesired anisotropy.

\subsection{Mutating genotypes}
Mutation event occurs in every simulation iteration. A small change in value is applied to a randomly choosen gene (random parameter type $p$, random kernel index $k$, random delta amount $\delta$) over a small region (random center position $(x_0, y_0)$, fixed square size $2s$) controlled by mutation rate $\gamma_\text{mut}$. Mutation only takes effect inside the alpha channel mask $\alpha$ (calculated in Section~\ref{seg:localize}).
\[P_{pkxy} \mapsto P_{pkxy} + \gamma_\text{mut} \, \delta \;\; \text{ where } (x,y) \in (x_0 \pm s, y_0 \pm s) \wedge \alpha\]

Due to the diffusion process in genospace (Section~\ref{sec:diffuse}), a dose of mutation will be diffused and averaged with its nearby connected area. The mutation may change the structure and behavior of the affected self-organizing pattern, or destabilize it into extinction or explosion.

\subsection{Penalize exploding patterns}
If a stable pattern is mutated into an exploding pattern, it quickly expands and occupies the nearby phenospace and genospace, potentially destroying other patterns in its way. There would be a whole set of interesting behaviors governing these non-local patterns that warrants more research (e.g. in MNCA \cite{Kraakman2021-mj}), but here we focus on the evolution of spatially localized enactivisitc agents. Moreover, exploding patterns tend to wide out all genetic diversity inside the simulation, therefore their existence is undesired in this work.

To detect and penalize exploding patterns, the basic idea is: imagine the empty space contains ``oxygen'' diffusing into organic matter down to a certain depth $r_\text{oxy}$, if a pattern grows too big, its inner parts will suffocate and die out. This idea is implemented by an inverse diffusion, i.e. convolve the empty space (reverse of organic matter) with an oxygen diffusion kernel $K_\text{oxy}$ to get the oxygenated area $E_\text{oxy}$, then convolve the unoxygenated area (reverse of $E_\text{oxy}$) with the same kernel to get the penalizing area $E_\text{pen}$. $K_\text{oxy}$ is a simple disk of radius $r_\text{oxy}$ pre-calculated during initialization.
\[E_\text{oxy} = (1-\Delta_\text{sum}(U))*K_\text{oxy}\]
\[E_\text{pen} = (1-E_\text{oxy})*K_\text{oxy}\]

\begin{figure}[htbp]
\centering
\includegraphics[width=0.8\linewidth]{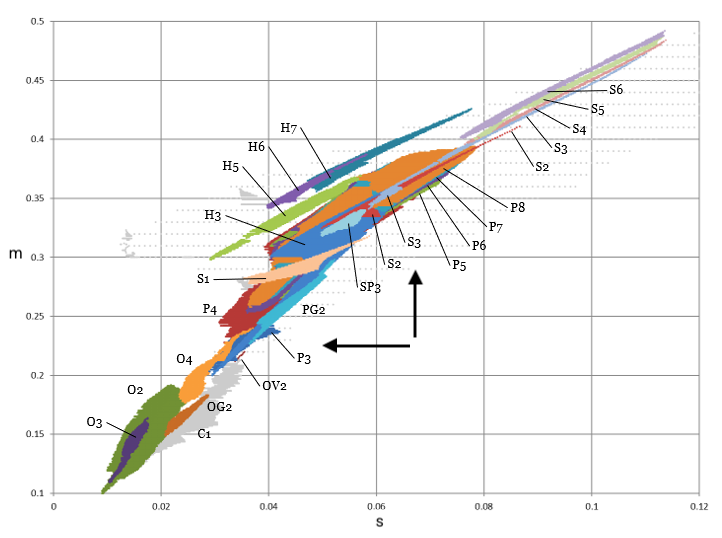}
\caption{$\mathsf m$-$\mathsf s$ parameter space of simple Lenia creatures, each color region represents the parameter range of a stable species. Arrows show how adjusting parameter values could stabilize an exploding pattern. Modified from \cite{Chan2019-po}.}
\label{fig:map} 
\end{figure}

There are several ways to penalize expanding patterns, e.g. to removed mass immediately or gradually. The algorithm used here is derived from the ``edge of chaos'' chart \cite{Chan2019-po} (Figure~\ref{fig:map}). If a pattern is exploding (lower right part), increase $\mathsf m$ or decrease $\mathsf s$ (arrows) would slow down the pattern growth and restore homeostasis (and \textit{vice versa} for vanishing patterns in the upper left part). Although not guaranteed to be effective, this principle was successfully used in manual search of numerous stable creatures \cite{Chan2019-po}. Here, a penalizing direction is randomly chosen (random choice of $p$ as $\mathsf m$ or $\mathsf s$, random kernel index $k$, random delta amount $\delta$ with its sign depends on $p$), controlled by penalization rate $\gamma_\text{pen}$.
\[P_{pkxy} \mapsto P_{pkxy} + \gamma_\text{pen} \, \delta \;\; \text{ where } (x,y) \in E_\text{pen} \wedge \alpha\]

A disadvantage is that if a swarm is formed by multiple localized creatures that are close together, they would be treated as a single area and incorrectly penalized by this algorithm.

\subsection{Environmental design}
Apart from the organic matter, environmental elements can be added to encourage desired behaviors, like in Flow Lenia \cite{Plantec2022-xw}. Here wall-like obstacles can be added by constantly removing materials inside the wall area in the phenospace and genospace. This effectively segment the world into connected rectangles, or a long narrow strip for better producing and protecting geographic speciation (cf. salamander evolution \cite{Moritz1992-zs})

\subsection{Interactive interface}
Following the tradition of interactive evolutionary computation (IEC), a user interface (UI) is designed for interacting with the simulation in real-time, to facilitate experimentation and iteration of ideas. UI panels display the phenospace, the genospace, kernel and growth functions, and mutation and penalization events. There are UI functions for restarting simulation, changing simulation speed, changing mutation and penalization rates, switching environmental elements on or off, inspecting a small area of the world with particular genotype and phenotype (``dropper'' tool).

\begin{figure}[htbp]
\centering
(a)\;\includegraphics[width=0.9\linewidth]{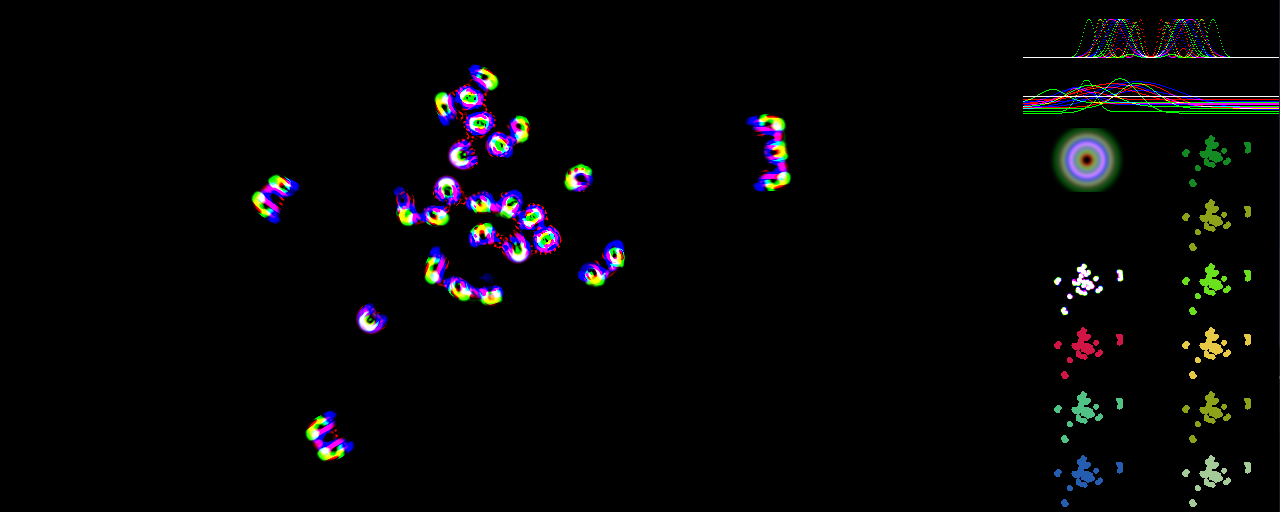}\\
(b)\;\includegraphics[width=0.9\linewidth]{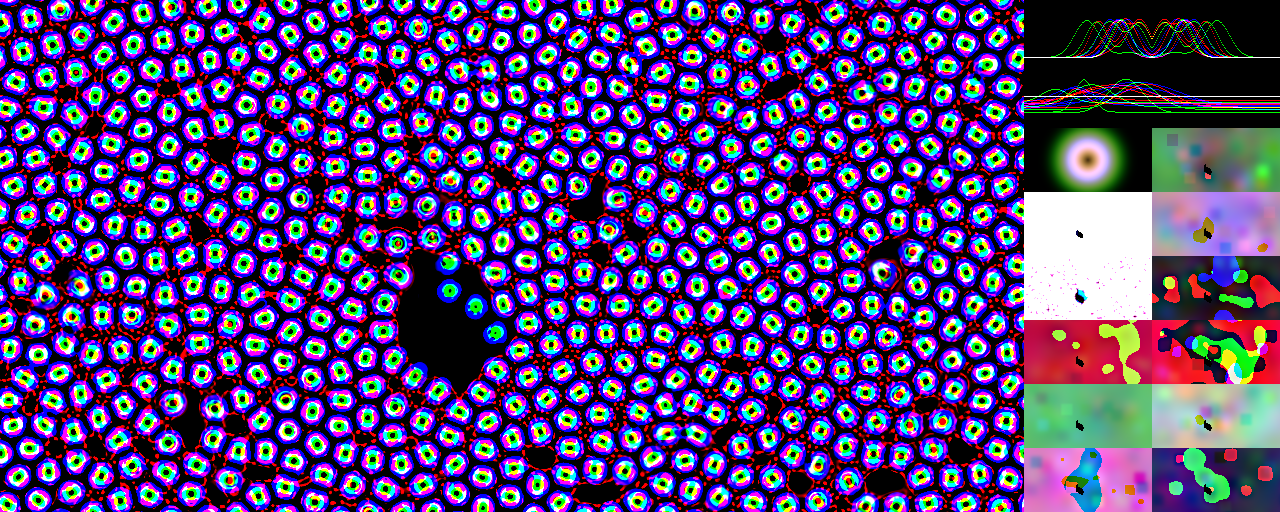}\\
(c)\;\includegraphics[width=0.9\linewidth]{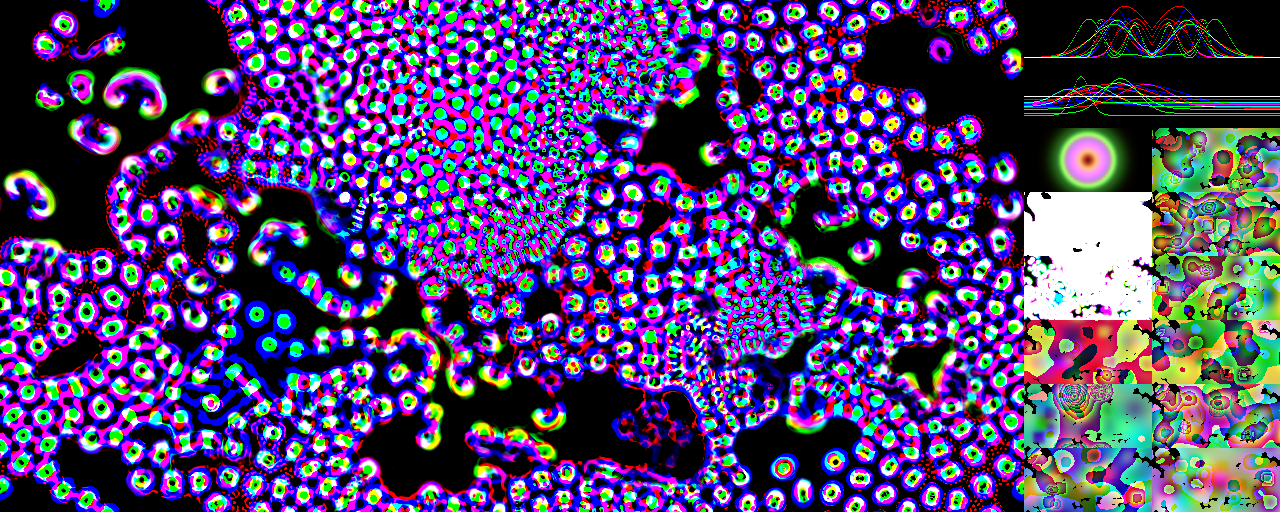}\\
(d)\;\includegraphics[width=0.9\linewidth]{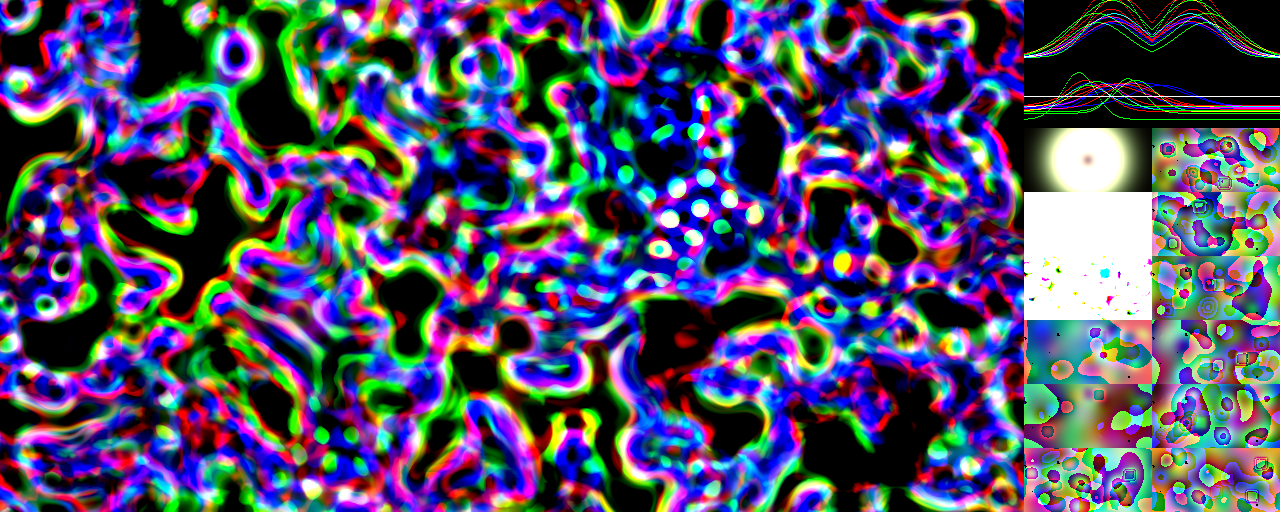}\\
(e)\;\includegraphics[width=0.9\linewidth]{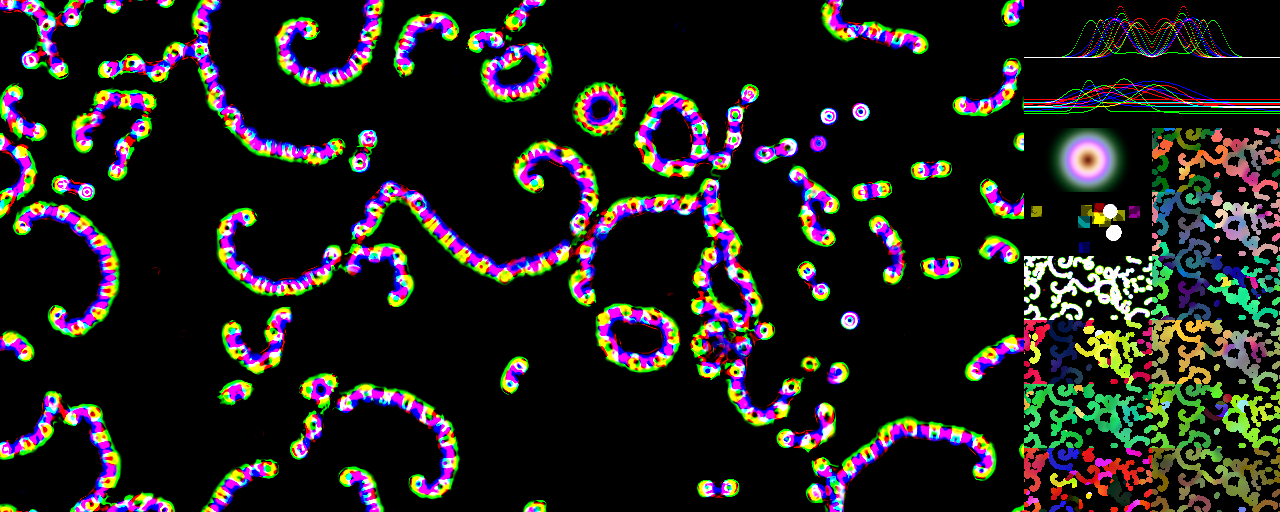}\\
(f)\;\includegraphics[width=0.9\linewidth]{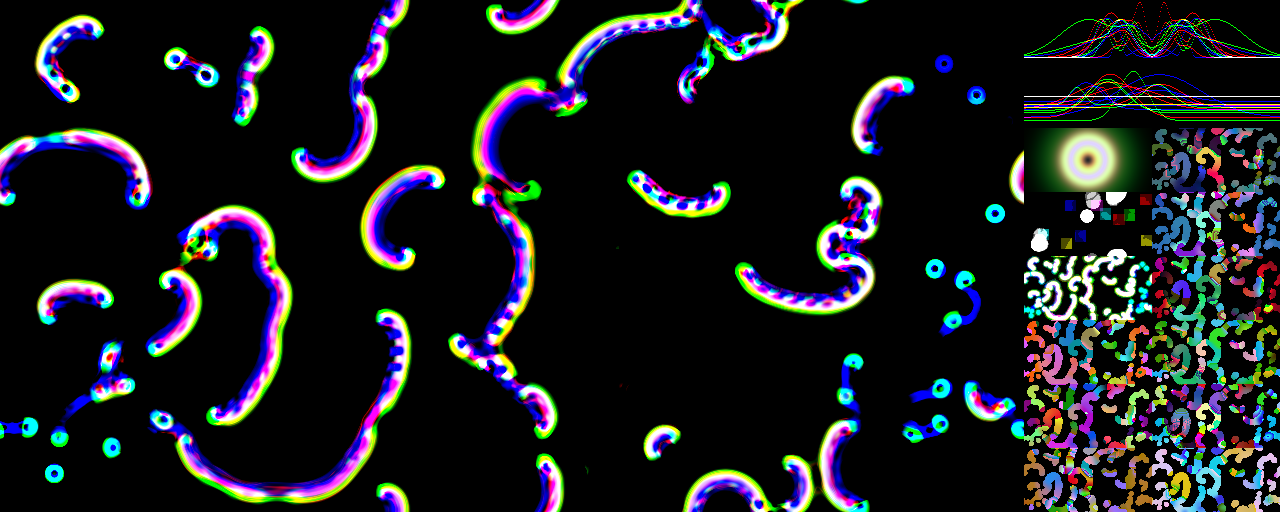}
\caption{Screenshots of large-scale evolutionary simulations. (a) Starting phase; (b) ending ``packed'' phase with no penalization + low mutation; (c, d) intermediate ``creative'' phase and ending ``goo'' phase with no penalization + high mutation; (e, f) ending ``linear'' phase with penalization + low or high mutation.}
\label{fig:screen} 
\end{figure}

\begin{figure}[htbp]
\centering
\includegraphics[width=0.24\linewidth]{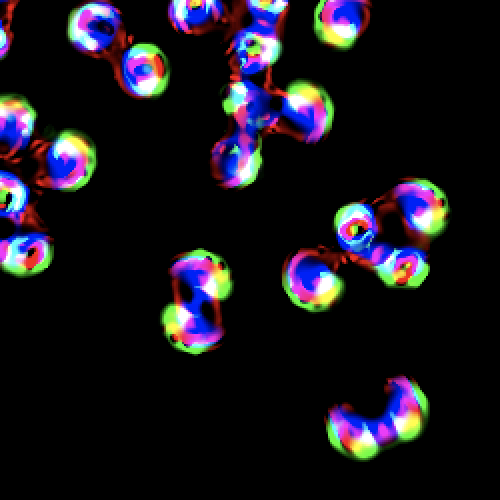}
\includegraphics[width=0.24\linewidth]{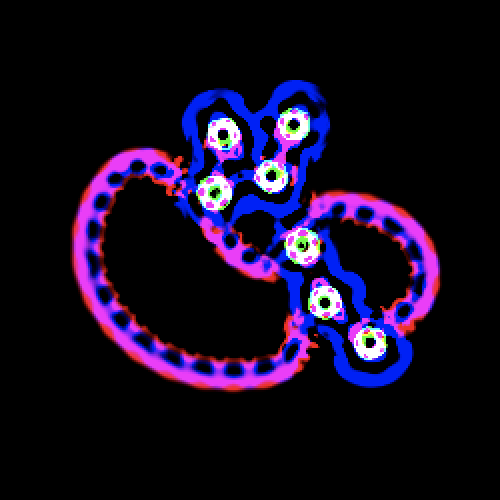} \includegraphics[width=0.24\linewidth]{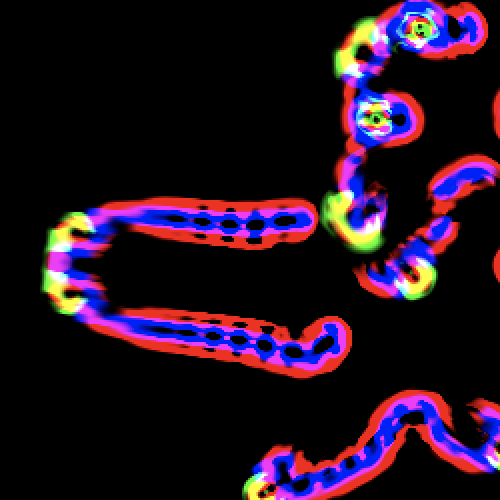}
\includegraphics[width=0.24\linewidth]{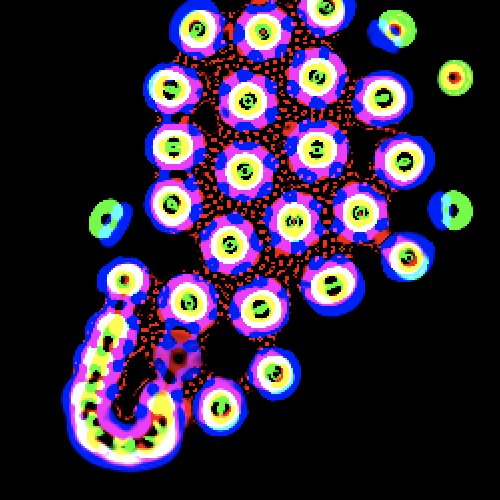} \\
\includegraphics[width=0.24\linewidth]{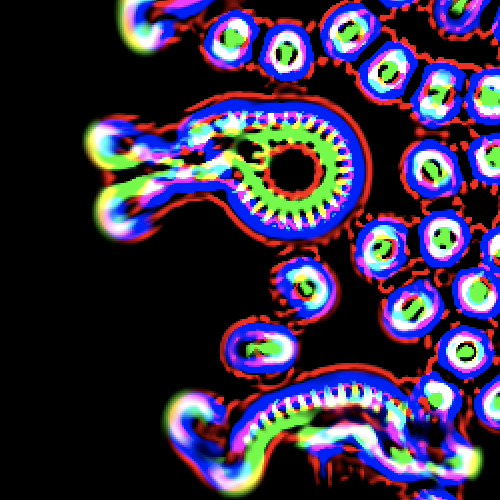}
\includegraphics[width=0.24\linewidth]{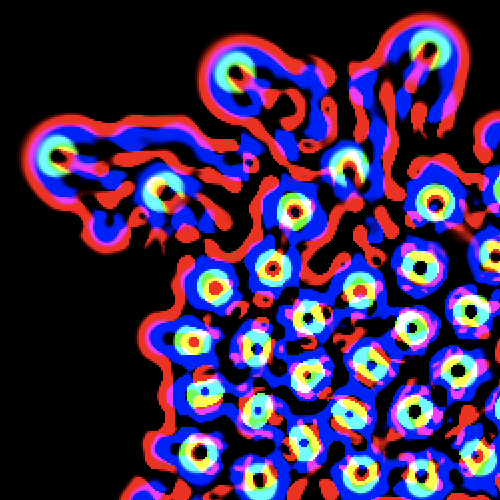} \includegraphics[width=0.24\linewidth]{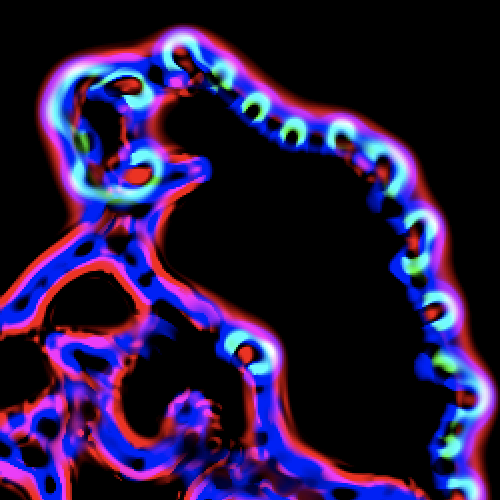}
\includegraphics[width=0.24\linewidth]{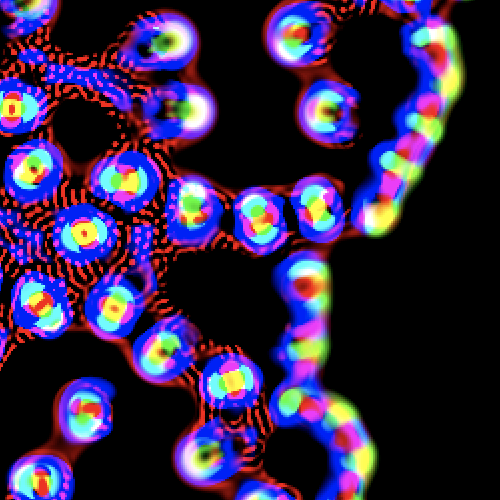} \\
\includegraphics[width=0.24\linewidth]{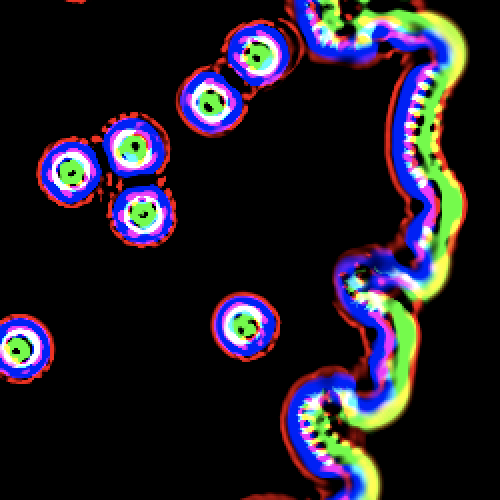}
\includegraphics[width=0.24\linewidth]{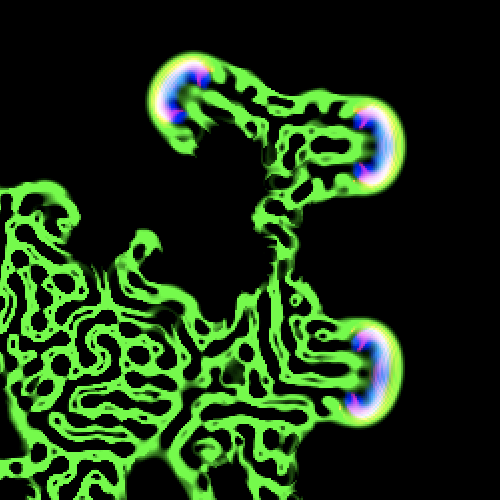} \includegraphics[width=0.24\linewidth]{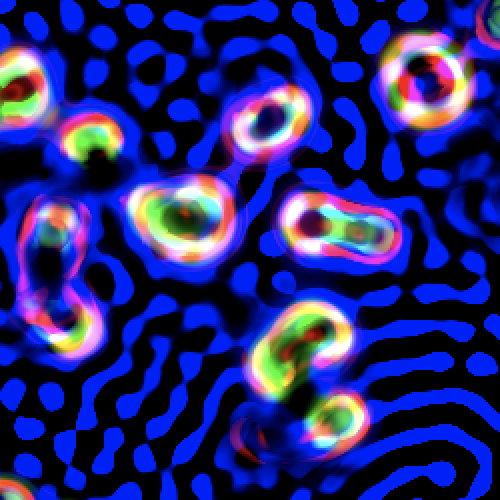}
\includegraphics[width=0.24\linewidth]{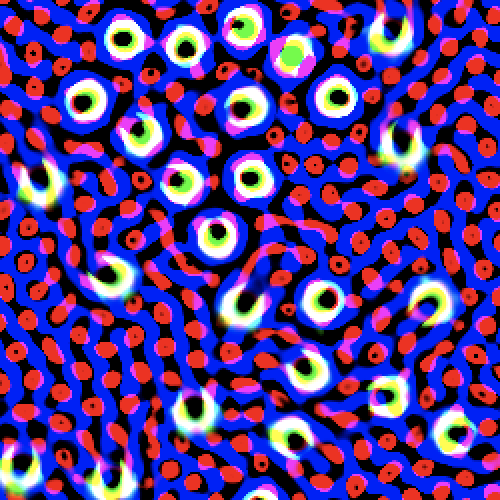} \\
\includegraphics[width=0.24\linewidth]{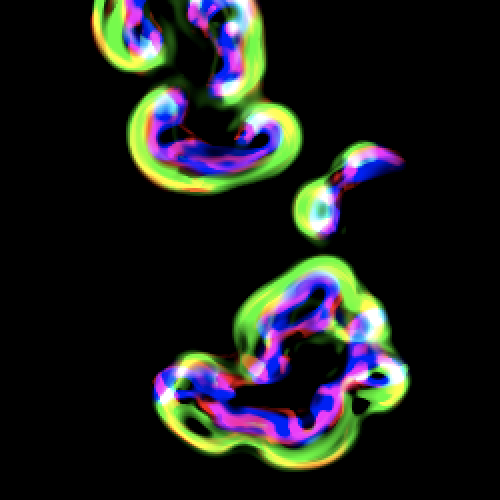}
\includegraphics[width=0.24\linewidth]{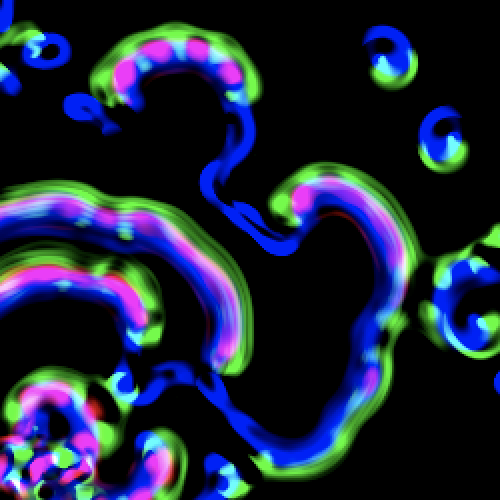}
\includegraphics[width=0.24\linewidth]{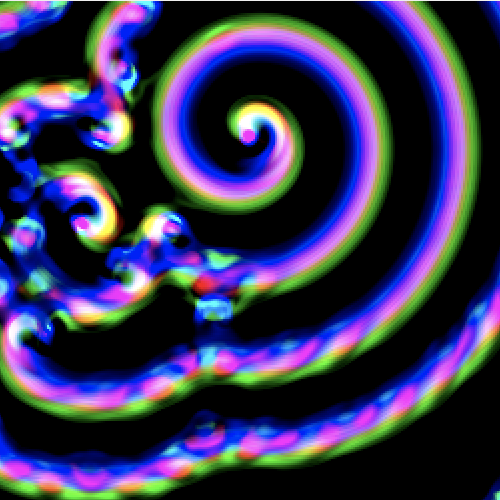}
\includegraphics[width=0.24\linewidth]{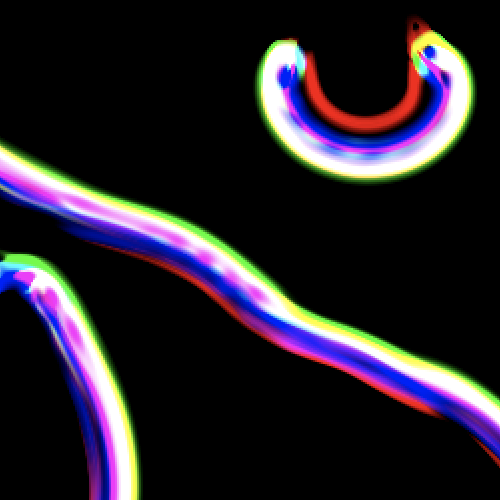}
\caption{Evolved patterns sampled from 5 evolutionary simulation runs, showing the creativity and diversity produced using high mutation rates. Each panel is part of a separate non-evolving simulation, with genotype and phenotype sampled from the main evolutionary simulations like Figure~\ref{fig:screen} (using the ``dropper'' tool).}
\label{fig:parts} 
\end{figure}

\section{Results}

Since we wish to replicate the evolution of the \textit{Aquarium} lineage as explained in the introduction, we use a self-replicating species early in the lineage (Figure~\ref{fig:lineage}, ``VT049W'') as the seed pattern. A few instances are randomly placed near the center, allowed to self-replicate for a period of time (Figure~\ref{fig:screen}a). Later, mutation and penalization can be switched on. Below describes the typical outcomes of 4 types of simulation runs with different penalization and mutation rates.

With penalization off and normal mutation rate ($\gamma_\text{pen} = 0.0, \gamma_\text{mut} = 1.0$), the world quickly becomes saturated by replicated entities, there is no sign of further evolution (Figure~\ref{fig:screen}b).

With penalization off and high mutation rate ($\gamma_\text{pen} = 0.0, \gamma_\text{mut} = 5.0$), the simulation goes through a transition phase producing diverse patterns with interesting global and local dynamics (Figure~\ref{fig:screen}c). The world effectively becomes a map or catalogue of those evolved patterns that can be inspected separately using the ``dropper'' tool  (Figure~\ref{fig:parts}) (cf. the parameter map in \cite{Munafo2014-er, Munafo2022-uu}). Finally the world converges into a kind of global ``goo'' patterns consists of violently flickering and fast moving \textit{quadratic expanding patterns} (QEPs), without differentiation into individual entities (Figure~\ref{fig:screen}d). This evolution result is apparently optimal for fast expansion of patterns.

With penalization on and normal mutation rate ($\gamma_\text{pen} = 0.2, \gamma_\text{mut} = 1.0$), the world produces interesting entities, mostly variants of the \textit{Aquarium} species, but are less diverse than the non-penalized version. Some \textit{linear expanding patterns} (LEPs) start to appear. Later the simulation may end towards total extinction, or evolves into domination by LEPs, which is apparently an optimal solution for this setting. LEPs manage to replace other forms of patterns and bypass the penalization mechanism (Figure~\ref{fig:screen}e).

With penalization on and high mutation rate ($\gamma_\text{pen} = 0.2, \gamma_\text{mut} = 5.0$), the world quickly converges into an optimal state, dominated by LEPs that are more robust (Figure~\ref{fig:screen}f).

In summary, either in the normal scenario (expanding patterns penalized) or the high intensity scenario (expanding patterns allowed but constantly bombarded by high mutation rate), the simulations go through a transition phase where diverse patterns evolved from the \textit{Aquarium} seed. Some of the evolved creatures have characteristics similar to the \textit{Aquarium} lineage successfully emerged or re-emerged (e.g. red ``communicating'' dots, self-replication). Eventually the world converges to either total extinction, or domination by expanding patterns, forbidding further progress into OEE.

\section{Discussions}

The experiments described in this work were considered partially successful, in the sense that periods of creativity and diversity before convergence have been observed, and the algorithm is able to find its optimal solutions given the implicit selection criteria. OEE has not been achieved, largely because of the dominance of expanding patterns like QEPs and LEPs that quickly destroy existing genetic diversity.

The follow sections discuss the lesson learned from these experiments and hints from the literature, and possible ways to further improve the evolutionary simulation.

\subsection{Ingredients of open-ended evolution}
We propose a number of factors that may facilitate successful OEE simulations in continuous CAs or other self-organizing systems.

In terms of simulation efficiency,
\begin{itemize}
    \item Better utilization of parallel computing on GPUs or TPUs to afford faster computation and larger world sizes; 
    \item Deriving algorithms that use multiple GPUs/TPUs in running a single simulation (enabled by the possibility of asynchronous update in continuous CAs); 
    \item Coarse-graining of the simulation for more efficiency; 
    \item Implementing software shortcuts for known structures and behaviors \cite{Banzhaf2016-kt}, instead of simulating all the way from the bottom up.
\end{itemize}

In terms of base system design, 
\begin{itemize}
    \item Localization of genotypes to enable intrinsic inheritance and species-species interactions, may use methods other than ring-wise kernels, e.g. weighted combination of ``casted'' kernels of various ranges (e.g. \cite{Plantec2022-xw}) -- this is more compute efficient but may limit the expressiveness of kernels;
    \item Decision of how genotype boundaries are treated, e.g. slow diffusion, hard separation, decide by flow strength in Flow Lenia (see Section~\ref{sec:flow} below) or by other competition functions \cite{Sayama2018-oc}.
    \item Mass conservation \cite{Plantec2022-xw, Mordvintsev2022-mb} with zero or controlled mass change, or other ways that can intrinsically eliminate expanding patterns like QEPs or LEPs without explicit penalization -- however, mass conservation may limit the creativity and expressiveness of the system; 
    \item Increase the degrees of freedom, e.g. more kernels, more channels, higher spatial dimensions, to give more room for morphological computation inside patterns.
\end{itemize}

In terms of evolutionary algorithm, 
\begin{itemize}
    \item Establishing localized virtual creatures as the units of selection;
    \item More genetic operators inspired by biology, e.g. sexual reproduction (mix genes i.e. parameters when two patterns collide and produce a third entity), gene duplication (duplicate a kernel-growth pair, allow new  functionalities while retaining the original ones);
    \item More forms of selection, e.g. differential existential success (i.e. robustness as well as competitiveness for space), differential reproductive success (i.e. effectiveness in self-replication or sexual reproduction);
    \item Energy constraints, e.g. creatures need to collect ``food'' or ``energy'' in order to survive, may serve as an incentive for more complex individual or collective behaviors;
    \item Virtual environmental design, e.g. obstacles, world segregation, energy sources, episodes of catastrophes, large area environment changes to simulate geographic speciation \cite{Sayama2018-oc}.
\end{itemize}

In terms of assessment, 
\begin{itemize}
    \item Sparse feedback from human-in-the-loop for selecting simulation episodes and tuning hyperparameters (cf. \cite{Etcheverry2020-yp});
    \item Detection and harvesting of individual creatures to assess the genetic and phenotypic diversity \cite{Sayama2018-oc};
    \item Quantitative measurements (e.g. entropy based) to assess the evolutionary dynamics \cite{Sayama2011-sp, Sayama2018-oc}.
\end{itemize}

\subsection{New kinds of emergence}
Provided that the right combination of the above-mentioned factor can facilitate further complexification or even OEE, we should look for any new form of emergence, for example,

\begin{itemize}
    \item New levels of organization, swarm entities that are capable of group-level collective behaviors, e.g. coordinated movements, autopoiesis (i.e. regeneration and self-replication) at group level, collective interactions;
    \item New forms of interaction, e.g. competition, cooperation, symbiosis, predation, parasitism;
    \item New forms of perception, that is, emission and reception of small signaling patterns, e.g. slow moving ``molecules'' for ``chemoreception'', or fast moving ``photons'' for ``vision'';
    \item New forms of inheritance, e.g. sexual reproduction, developmental biology;
    \item New kinds of computation, e.g. memory, learning, logic gates
\end{itemize}

Many of these emergent phenomena are already possible given the previous observations. Examples include group-level swarming, exchange of small red signalling dots, budding and emission of small red gliders (Figure~\ref{fig:lineage}), preliminary cases resembling predation and developmental biology, training of sensorimotor agency using gradient descent and curriculum learning \cite{Hamon2022-od}.

Note that observation of a new organization level or other more complex behaviors would require even bigger worlds and longer timescales \cite{Banzhaf2016-kt}, as more space-time would be needed for more complex morphological computation, and for the expansion of the virtual creatures' cognitive horizons \cite{Levin2019-wi}.

\subsection{Limitations}
There are a few possible limitation in achieving OEE. Artificial life systems are difficult to scale due to computational requirement or large-scale simulations. In the case of continuous CAs, it is limited by the speed of FFT, i.e. $\mathscr O(N \log N)$ where $N$ is the number of total pixels in the world.

There may be a plateau of evolutionary creativity even in biological life and technological advances due to physical limitations. For example, increments in the Sepkoski curve (showing biodiversity through geological time) and the Moore's law (showing  transistor counts in microprocessors) may actually be slowing down.

\subsection{Similar attempts} \label{sec:flow}
\begin{figure}[htbp]
\centering
(a)\;\includegraphics[width=0.4\linewidth]{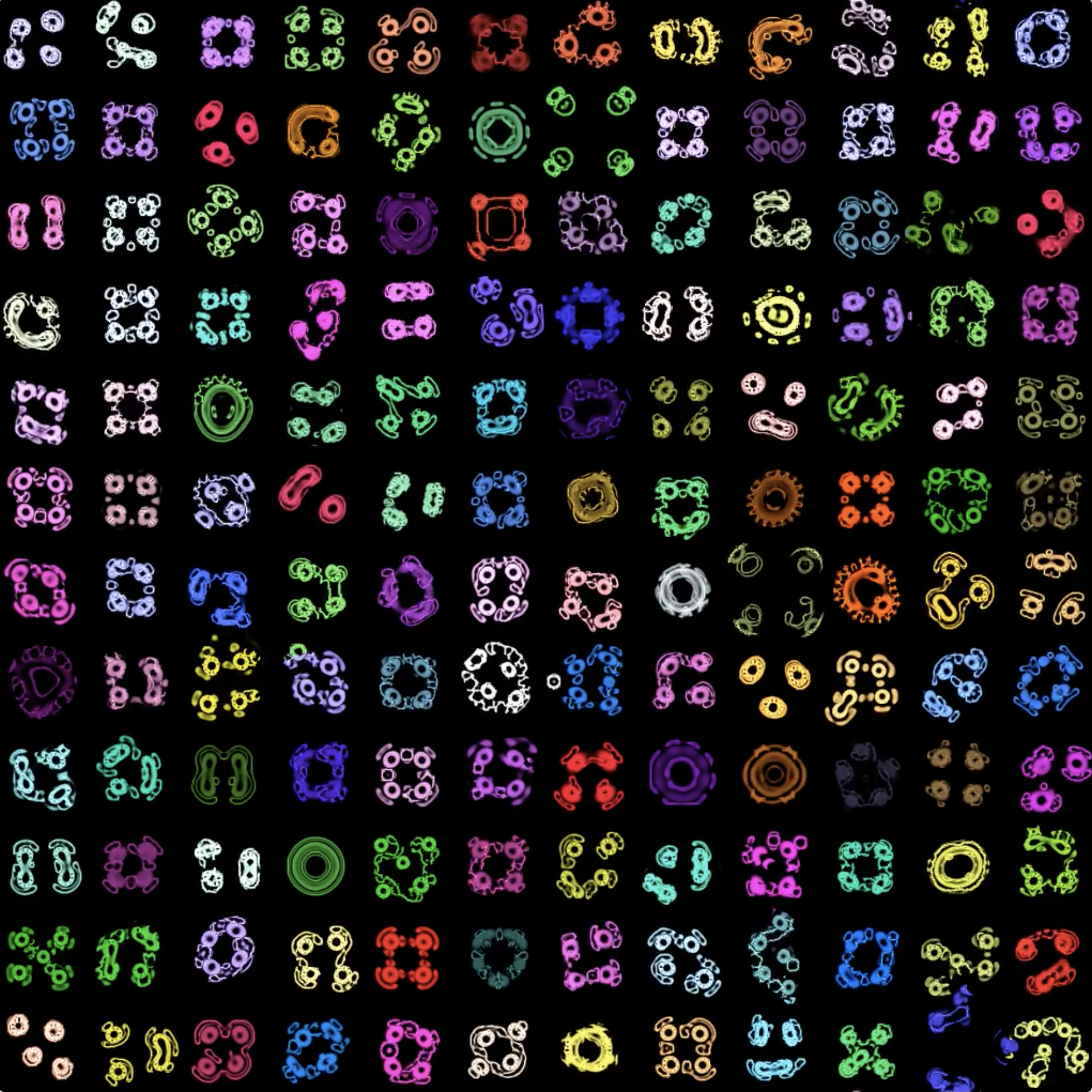}
\includegraphics[width=0.4\linewidth]{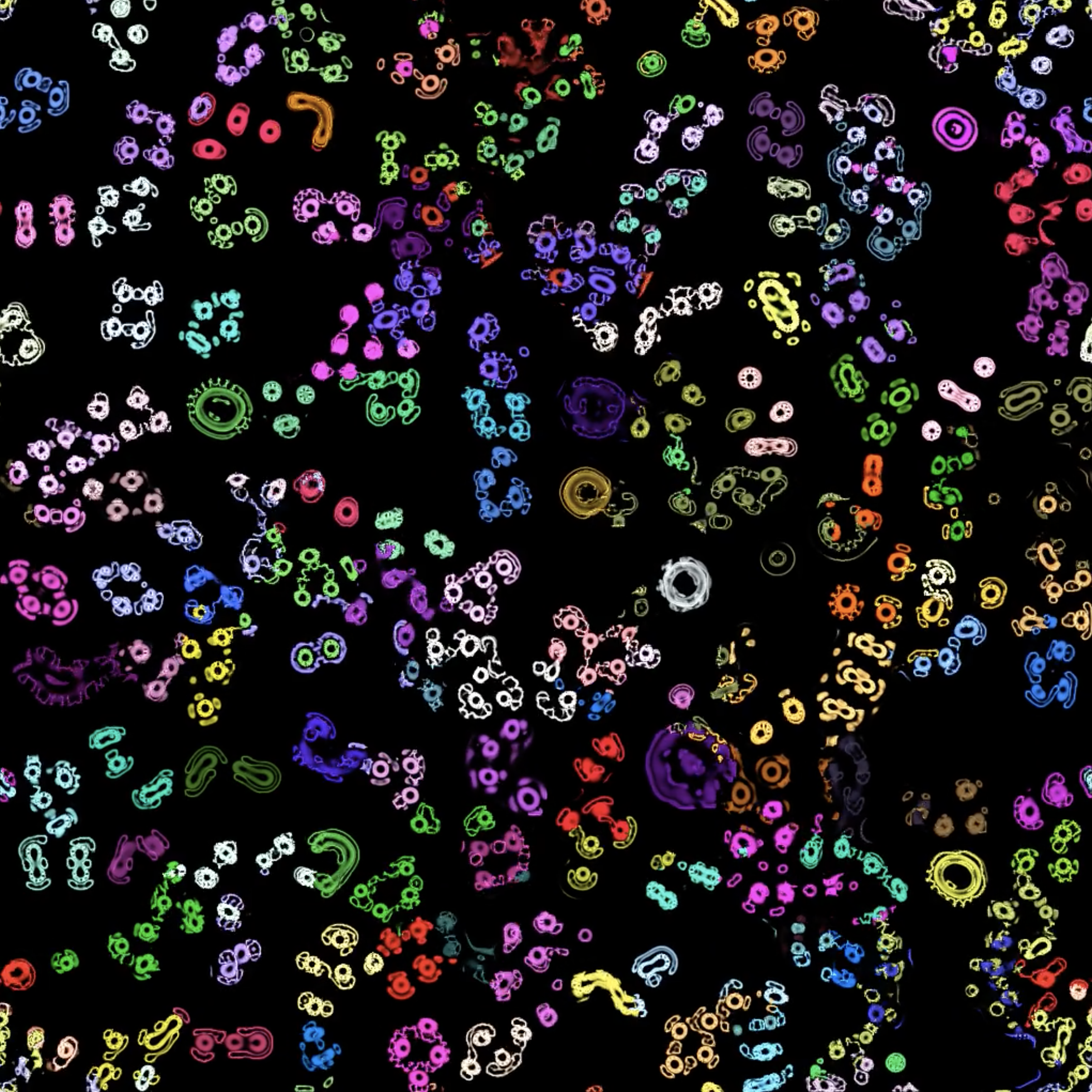}\;(b) \\
(c)\;\includegraphics[width=0.4\linewidth]{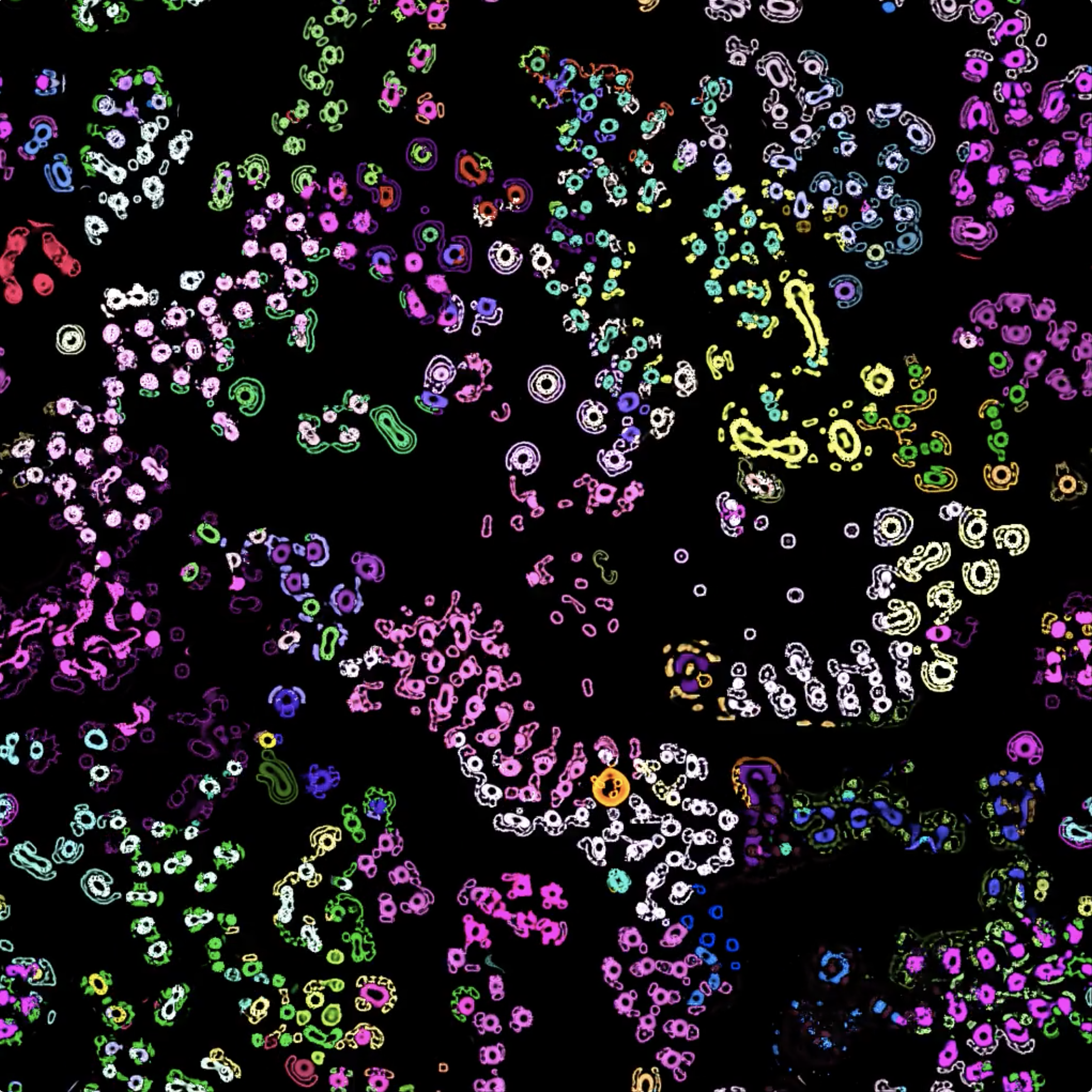}
\includegraphics[width=0.4\linewidth]{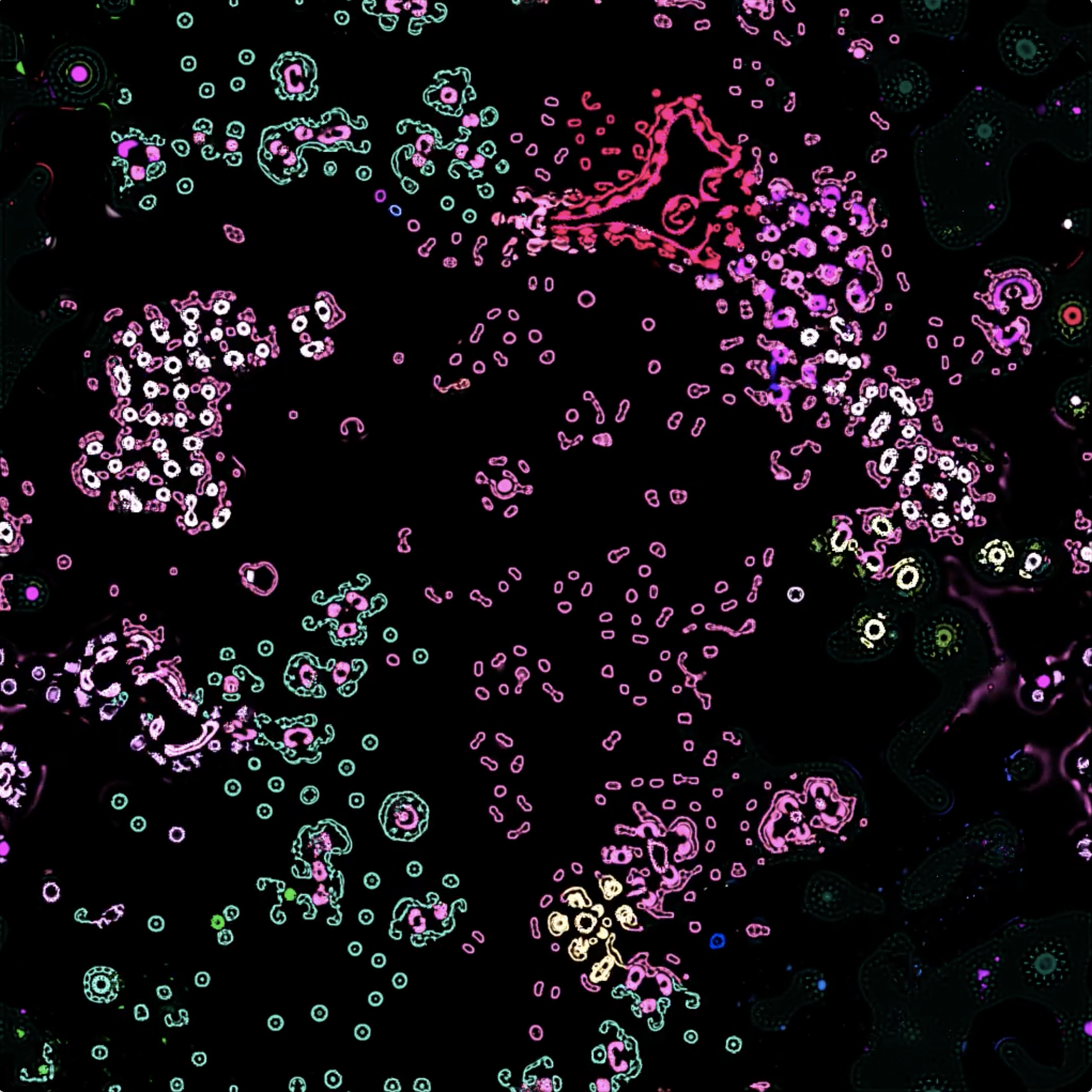}\;(d)
\caption{Large-scale evolutionary simulation using Flow Lenia with mass conservation. (a) Starting with 144 species; (b) competition for space; (c, d) domination of a few species.}
\label{fig:flow} 
\end{figure}

Using Flow Lenia \cite{Plantec2022-xw}, a variant of Lenia with mass conservation mechanism, similar large-scale evolutionary simulations have been run (Figure~\ref{fig:flow}) [Erwan Plantec, unpublished data]. The simulation has a world size of $1024 \times 1024$, and is initialized with 144 species of different genotypes. During the run, the species compete for space, with some of them start to win over, and species-species symbiosis seems to occur. In later stage, the world is dominated by a few species and multi-species alliances. The apparent species-level symbiotic and competitive relationships are particularly interesting.

\begin{acks}
We would like to thank Erwan Plantec for providing the experiment results in section \ref{sec:flow}; Gautier Hamon, Mayalen Etcheverry, Clément Moulin-Frier, Pierre-Yves Oudeyer, Yingtao Tian and Yujin Tang for discussions and inspirations; Kenneth Stanley, Joel Lehman, Jeff Clune and Lisa Soros for advocating research on open-endedness.
\end{acks}

\bibliographystyle{ACM-Reference-Format}
\bibliography{chan2023}

\appendix

\end{document}